\documentclass[a4paper,11pt]{article}
\usepackage[utf8]{inputenc}

\usepackage[scaled]{helvet}
\usepackage[T1]{fontenc}

\RequirePackage{graphicx}
\usepackage[usenames,dvipsnames]{xcolor}
\usepackage[scale=0.8]{geometry}
\usepackage{float}

\usepackage[unicode, draft=false]{hyperref}
\definecolor{linkcolour}{rgb}{0,0.2,0.6}
\hypersetup{colorlinks,breaklinks,citecolor=linkcolour, urlcolor=linkcolour,linkcolor=linkcolour}

\usepackage[authoryear]{natbib}
\usepackage{mathrsfs,amssymb}
\usepackage{amsmath,color}
\usepackage{hyperref}

\usepackage{mathtools}
\usepackage[all]{xy}
\newcommand{\bcx}{{\bf X}}

\newcommand{\bcz}{{\bf Z}}

\newcommand{\bc}{{\bf c}}
\newcommand{\bx}{{\bf x}}

\newcommand{\bz}{{\bf z}}

\newcommand{\bt}{{\bf t}}
\newcommand{\bmu}{{\boldsymbol\mu}}
\newcommand{\bfphi}{{\boldsymbol\varphi}}
\newcommand{\bfeta}{{\boldsymbol\eta}}

\newcommand{\btheta}{{\boldsymbol\theta}}

\newcommand{\bfsigma}{{\boldsymbol\sigma}}

\newcommand{\R}{\mathbb{R}}
\begin{document}
	
\title{Investigating a domain adaptation approach for integrating different measurement instruments in a longitudinal clinical registry}

\author{Maren Hackenberg$^{*,1,2}$, 
        Michelle Pfaffenlehner$^{1,2}$, 
        Max Behrens$^{1,2}$, \\ 
        Astrid Pechmann$^{3}$, 
        Janbernd Kirschner$^3$, and 
        Harald Binder$^{1,2,4}$
}

\date{}

\maketitle 

\noindent $^1$ Institute of Medical Biometry and Statistics, Faculty of Medicine and Medical Center, University of Freiburg, Freiburg, Germany \\
$^2$ Freiburg Center for Data Analysis and Modeling, University of Freiburg, Freiburg, Germany \\
$^3$ Department of Neuropediatrics and Muscle Disorders, Faculty of Medicine and Medical Center, University of Freiburg, Freiburg, Germany \\
$^4$ Centre for Integrative Biological Signalling Studies (CIBSS), University of Freiburg, Freiburg, Germany \\ 
$^*$ Corresponding author: {\sf{maren.hackenberg@uniklinik-freiburg.de}} \\

\begin{abstract}
In a longitudinal clinical registry, different measurement instruments might have been used for assessing individuals at different time points. To combine them, we investigate deep learning techniques for obtaining a joint latent representation, to which the items of different measurement instruments are mapped. This corresponds to domain adaptation, an established concept in computer science for image data. Using the proposed approach as an example, we evaluate the potential of domain adaptation in a longitudinal cohort setting with a rather small number of time points, motivated by an application with different motor function measurement instruments in a registry of spinal muscular atrophy (SMA) patients. There, we model trajectories in the latent representation by ordinary differential equations (ODEs), where person-specific ODE parameters are inferred from baseline characteristics. The goodness of fit and complexity of the ODE solutions then allows to judge the measurement instrument mappings. We subsequently explore how alignment can be improved by incorporating corresponding penalty terms into model fitting. To systematically investigate the effect of differences between measurement instruments, we consider several scenarios based on modified SMA data, including scenarios where a mapping should be feasible in principle and scenarios where no perfect mapping is available. While misalignment increases in more complex scenarios, some structure is still recovered, even if the availability of measurement instruments depends on patient state. A reasonable mapping is feasible also in the more complex real SMA dataset.
These results indicate that domain adaptation might be more generally useful in statistical modeling for longitudinal registry data.
\end{abstract}


\section{Introduction}
\label{sec:intro}

In longitudinal data collection in a clinical registry, often different measurement instruments are used over time to assess individuals, e.g., as some instruments might be more appropriate for individuals of a certain age, or some might be used only sporadically. To obtain more information on an individual's development for modeling, it would thus be beneficial to integrate these instruments. For high-volume data, such as images, domain adaptation approaches have been quite successful in mapping data from different domains, such as different modes of microscopy, into a joint latent representation \citep{Long2015, Tzeng2017}. Therefore, we want investigate the potential of such a domain adaptation approach in the setting of clinical registries, with a particular focus on setting with a small number of time points. 
This is motivated by our work with the SMArtCARE registry \citep{Pechmann2019} comprising patients with spinal muscular atrophy (SMA). There, different physiotherapeutic assessments, each comprising a larger number of motor function characteristics as items, are available for each of the affected patients over time, reflecting individual disease development. Due to limited time for assessment, a patient's age, and disease severity, different physiotheurapeutic tests have been used at different time points for capturing different aspects of motor function. For example, the Hammersmith Functional Motor Scale Expanded (HFMSE) test \citep{Glanzman2011, OHagen2007} assesses gross motor function skills while the Revised Upper Limb Module (RULM) test \citep{Mazzone2017} is used to assess fine motor function skills of the upper limbs. 
While both tests are informative about the underlying SMA disease development, they may capture some overlapping, but also some distinct aspects. 
Domain adaptation then means to map the items of these different measurement instruments to a low-dimensional joint latent representation. 

In computer science, domain adaptation is a well-established concept, and a large number of approaches have been proposed \citep[for a recent overview, see, e.g.,][]{Farahani2021}. 
These address the challenge of linking knowledge between data from a source and a target domain, based on the assumption that an underlying latent concept is shared between the domains, and that it is feasible to map both domains into a joint representation. Typically, deep learning approaches are used to learn such domain-invariant representations \citep[e.g.,][]{Chen2012, Long2015, Tzeng2017}. 
So far, many approaches focus on image data \citep[e.g.,][]{Csurka2017, Ganin2015} and have been successfully applied also in the biomedical domain \citep[for an overview, see][]{Guan2022}, e.g., for MR imaging \citep[e.g.,][]{Goetz2016, Ghafoorian2017, Karani2018} or microscopy \citep[e.g.,][]{Becker2015}. 
Recently, similar approaches have also been adapted to high-frequency time series data \citep[e.g.,][]{Jin2022, He2023, Lai2023}, including biomedical applications \citep[e.g.,][]{Purushotham2016, Jude2022, Raghu2023}. As these approaches focus on rather large datasets with many individuals and time points per individual, they cannot be directly adapted for typical clinical registry settings with a potentially small number of patients, such as for a rare disease, and only few time points per patient. 

For our investigation, we therefore propose an approach tailored to this setting. It is based on variational autoencoders (VAEs) \citep{Kingma2014, Kingma2019}, which use artificial neural networks for learning an encoder, i.e., a mapping to a latent representation, such that the most important features of the data can be reconstructed based on this representation. We use separate encoders for mapping the items of each measurement instrument, at each time point where it is available, to a joint latent representation, corresponding to applying different transformations to the measurement instruments. 
The joint representation then allows to model an individual's dynamics. We focus on modeling trajectories by ordinary differential equations (ODEs), to reflect growth dynamics and inhibition by disease. 
% In the latent space, as a criterion for a good alignment, we have fitted a joint ODE to the latent representations of both measurement instruments, 
The ODE solutions, fitted in the joint representation, then allow to assess how closely the representation of each measurement instrument aligns to the common trajectory. Specifically, we allow for patient-specific trajectories by inferring individual ODE parameters from a patient's baseline characteristics with an additional neural network, as these characteristics are assumed to be informative about individuals' disease dynamics \citep{HacHarPfa2022}. 
To encourage alignment in the latent space, we employ an additional penalty term as a regularization based on the differences of the representations of individual measurement instruments to the common ODE solution. This corresponds to the concept of adversarial domain adaptation \citep{Tzeng2017}, where a model is trained such that an adversarial classifier is unable to distinguish between data from different domains, i.e., different measurement instruments, in the latent representation, inspired by generative adversarial networks \citep{Goodfellow2014}.

Still, it may not be clear whether the measurement instruments can actually be perfectly aligned in a latent representation or whether there are systematic discrepancies in the measured information. For assessing this challenge, we draw inspiration from the rich statistical literature on measurement error (see \citet{Crainiceanu2006} for an overview). In particular, the problem of obtaining equivalent information from different instruments measuring the same constructs has been addressed by latent variable models \citep[e.g.,][]{Bauer2009, Buuren2005, Heuvel2020}, which often require anchor items present across different instruments. Also, latent class models have been employed for describing disease progression based on multivariate longitudinal data \citep{ProustLima2014, ProustLima2023}. 

For evaluating agreement between single score variables, simple measures can be used \citep{Bland1990}. To address potentially more complex settings in our evaluation, 
% studied in the context of data harmonization \citep[e.g.,][]{Fortier2010, Fortier2017, Griffith2013}, which often relies on a common set of core variables. 
we specifically want to consider settings with different degrees of overlap in the information of the two measurement instruments, such as shifts in one instrument for a subgroup of patients, or systematic differences in the time intervals where each measurement instrument has been used. For this, we have designed several scenarios based on artificially modified data from one measurement instrument in the SMArtCARE registry. This includes scenarios where a perfect mapping is available in principle and scenarios where this is not feasible. 

We proceed by giving a brief overview of VAEs for learning a latent representation for each measurement instrument 
%in Section \ref{subsec:methods-VAE}  
and describe our approach for combining VAEs with ODEs, for obtaining a joint latent trajectory and aligning the representations of different measurement instruments, in Section \ref{sec:methods}. %\ref{subsec:methods-ODE}. 
%Then, we explain how the model is regularized to encourage a close alignment.
%in Section \ref{subsec:methods-loss}. 
In Section \ref{sec:eval}, we empirically evaluate the potential of such a domain adaptation approach, based on the different scenarios described above.  %in Section \ref{subsec:eval-modifications} 
%in Section \ref{subsec:eval-modifications-results}. 
We further illustrate the approach on data from two SMArtCARE measurement instruments %in Section \ref{subsec:eval-SMArtCARE}, 
and discuss our findings in Section \ref{sec:discussion}. 

%--------------------------

%We consider dataset modifications to investigate the feasibility of a domain adaptation approach in different data scenarios with systematic discrepancies between measurement instruments. 
%by, e.g., considering a certain subscale only in a specific time interval, or adding a shift to a subscale for a subgroup of patients, reflecting the assumption of measurement error in this subscale due to patient-specific characteristics
% and evaluate to what extent it is possible to recover the full information of the measurement instrument based on the modified subscales using our approach
% investigate the effect of the synthetic modifications

% In the latent space, as a criterion for a good alignment, we have fitted a joint ODE to the latent representations of both measurement instruments, 

%--------------------------

\section{Methods}
\label{sec:methods}

%In the following, we describe our domain adaptation approach for mapping data from different measurement instruments into a joint latent space, where they are aligned. 
%We first introduce a model for learning a flexible low-dimensional representation, the variational autoencoder (VAE).  We then explain how separate VAEs, one for each measurement instrument, are used to map them to a latent space, and how these latent spaces can be aligned by fitting an ordinary differential equation (ODE) jointly to the representations of both measurement instruments. 
%To encourage a good mapping, we propose an additional penalty term. 

\subsection{VAEs for a per-measurement-instrument mapping to a latent space}
\label{subsec:methods-VAE}

To infer an underlying latent representation from the observed data of the items of each measurement instrument, we employ a VAE, a generative deep learning approach that uses artificial neural networks to learn a low-dimensional representation \citep{Kingma2014, Kingma2019}. 
Specifically, we use a separate VAE for each measurement instrument to map it to a latent representation that is shared between instruments.  
The VAE latent space is defined as a low-dimensional random variable $\bcz$ with prior distribution $P^{\bcz}$. Two distinctly parameterized neural networks, called the encoder and the decoder, map an observation $\bx \in \R^{p}$, i.e., the items of a measurement instrument for an individual at a specific time point, to the latent space and back to data space. The parameters of the encoder and decoder are jointly optimized to infer a low-dimensional representation that allows to reconstruct the original data well.
%We denote densities of random variables by small letters with the random variable as subscript, and abbreviate conditional distributions and densities by writing, e.g., $p_{\bcx\mid \bz}(\bx)$ for $p_{\bcx \mid \bcz =\bz}(\bx)$. 
Formally, the encoder and decoder parameterize the conditional distributions $q_{\bcz \mid \bx}(\cdot,\bfphi)$ and $p_{\bcx \mid \bz}(\cdot,\btheta)$, where we abbreviate conditional distributions and densities by writing, e.g., $p_{\bcx\mid \bz}(\bx, \btheta)$ for $p_{\bcx \mid \bcz =\bz}(\bx, \btheta)$. The model is trained, i.e., the parameters $\btheta$ and $\bfphi$ of the encoder and decoder networks are optimized, by maximizing the evidence lower bound (ELBO). The ELBO is a lower bound on the data likelihood $p_{\bcx}$, derived based on variational inference \citep{Blei2017}:
    \begin{equation*}
        \mathrm{ELBO}(\bx, \bfphi, \btheta) = E_{q_{\bcz \mid \bx}(\cdot, \bfphi)}[\log(p_{\bcx \mid \bz}(\bx, \btheta))] - D_{\mathrm{KL}}(q_{\bcz \mid \bx}(\cdot, \bfphi)\Vert p_{\bcz}) \leq \log(p_{\bcx}).
    \end{equation*}
% is motivated by

It can be shown that maximizing the ELBO is equivalent to minimizing the Kullback-Leibler divergence between the true but intractable posterior $p_{\bcz \mid \bx}(\cdot,\btheta)$ and its approximation by a member of a parametric variational family $\lbrace q_{\bcz \mid \bx}(\cdot,\bfphi) \rbrace$, typically assumed as a Gaussian with diagonal covariance matrix \citep{Blei2017, Kingma2014, Kingma2019}. 
The first term of the ELBO can be interpreted as a reconstruction error, while the second term acts as a regularizer that encourages densities close to the standard normal prior.
Training a VAE then corresponds to maximizing the ELBO as a function of the parameters $\btheta$ and $\bfphi$ of the encoder and decoder, specifying the conditional distributions, which yields both an approximate maximum likelihood estimate for $\btheta$ and an optimal variational density $q_{\bcz \mid \bx}(\cdot, \bfphi)$. Such optimization is typically performed by stochastic gradient descent \citep{Kingma2015}, where the so-called reparameterization trick is used to obtain gradients of the ELBO w.r.t. the variational parameters $\bfphi$ \citep{Kingma2014, Kingma2019}. 

To map data from two different measurement instruments into a joint space, we use separate VAEs for each of them and align their latent spaces. Specifically, for two measurement instruments $R$ and $S$, we represent the observed longitudinal data of the $i$th individual's measurement items as matrices $\bx^R_i \in \R^{p \times (T^R_i+1) }$, $\bx^S_i \in \R^{q \times (T^S_i+1)}$, of measurements of $p$ and $q$ items, respectively, at $T^R_i+1$ and $T^S_i+1$ time points, respectively. We define the individual-specific and measurement instrument-specific time points as $(\bt^{R}_i) := (t_0, t^{R,i}_1..., t^{R,i}_{T^R_i})$ and $(\bt^{S}_i) := (t_0, t^{S,i}_1..., t^{S,i}_{T^S_i})$, where $t_0$ denotes the common baseline time point. 
We define corresponding separate VAEs, $\mathrm{VAE}_R$ with parameters $\btheta_R, \bfphi_R$ and $\mathrm{VAE}_S$ with parameters $\btheta_S, \bfphi_S$, and for each latent space specify the variational posterior as multivariate Gaussian with diagonal covariance matrix, parameterized by the respective VAE's encoder. 

The encoder of $\mathrm{VAE}_R$ then maps the observed time series to the posterior mean $\bmu^R_i:= (\bmu_i^{R, t})_{t\in (\bt_i^R)} \in \R^{d\times (T^R_i+1)}$ and standard deviation $\bfsigma^R_i := (\bfsigma_i^{R, t})_{t \in (\bt_i^R)}\in \R^{d\times (T^R_i+1)}$, where $d$ is the dimension of the latent space. For the measurement instrument $S$, the posterior mean $\bmu^S_i$ and standard deviation $\bfsigma^S_i$, obtained from the encoder of $\mathrm{VAE}_S$, are defined analogously. 

\subsection{Aligning the latent spaces of each measurement instrument by fitting a joint trajectory via ODEs}
\label{subsec:methods-ODE}

To align the two latent spaces of $\mathrm{VAE}_R$ and $\mathrm{VAE}_S$, we solve a differential equation, using the latent representations from both measurement instruments to fit a joint trajectory. We then evaluate the joint solution at the time points of each measurement instruments and use the values of the joint trajectory at the time points of $R$ as input for the decoder of $\mathrm{VAE}_R$, and at the time points of $S$ for the decoder of $\mathrm{VAE}_S$. 

For obtaining the joint ODE solution, we build on an approach that we have proposed previously \citep{Hackenberg2023}. Specifically, we solve a linear ODE system multiple times, using each latent value obtained from the items of both measurement instruments observed in the course of time as the initial condition. The resulting individual solutions are combined into an inverse-variance weighted average using time-dependent weights, to obtain an unbiased minimum variance estimator of the true underlying dynamics \citep{Shahar2017}. This allows to reduce dependence of the ODE solution on the initial condition, i.e., the first value observed in the course of time, and incorporate all observed values into the fitting process. 
Specifically, we assume that for each individual $i$ the posterior means $\bmu_i^R, \bmu_i^S$ of the latent representations of both measurement instruments are governed by a common underlying $d$-dimensional linear ODE system with constant coefficients $A_i, \bc_i$. We drop the index $i$ for ease of notation and consider for a specific individual and for each $k=t_0, t^R_1, \dots, t^R_{T^R}, t^S_1, \dots, t^S_{T_S}$ the initial value problem 

\begin{align}
    \begin{split}
    \label{eq:ivp}
        \frac{d}{dt} \bmu(t) &= A\cdot \bmu(t) + \bc; \\
        \bmu(k) &= \bmu^{\lbrace R, S \rbrace, k} := \begin{cases} \bmu^{R,k}; & k \in (\bt^R), \\ \bmu^{S,k}; & k\in (\bt^S), \end{cases}
    \end{split}
\end{align}
where the latent mean at the $k$-th time point is used as the initial condition. 
For $A=0$, the solution to \eqref{eq:ivp} is given by $\bmu(t) = \bc\cdot t + \bmu^{\lbrace R, S \rbrace,k}$.
For $A \neq 0$, the solution can be computed analytically (see \citep{Hackenberg2023} for details):
\begin{align}
    \label{eq:analyticalsolution}
    \bmu(t) =\exp(A(t- k))\cdot(A^{-1}\bc + \bmu^{\lbrace R, S \rbrace,k}) - A^{-1}\bc.
\end{align}

We define the solutions of Equation \eqref{eq:ivp} for each $k=t_0, t^R_1, \dots, t^R_{T^R}, t^S_1, \dots, t^S_{T_S}$ as 
\begin{align*}
    \widetilde{\bmu}_k(t, \bfeta) := \begin{cases} (A^{-1}\bc + \bmu^{\lbrace R, S \rbrace,k}) \exp(A(t- k)) - A^{-1}\bc; & A \neq 0 \\ \bc\cdot t + \bmu^{\lbrace R, S \rbrace,k}; & A = 0, \end{cases}
\end{align*} 
where $\bfeta = \lbrace A, \bc \rbrace$ and $\bmu^{\lbrace R, S \rbrace,k}$ is used as initial condition. The solutions are then combined to a time-dependent inverse-variance weighted average as an estimator for the true underlying dynamics (see \citep{Hackenberg2023} for details):
\begin{align}
\label{eq:ODE_estimator}
    \widetilde{\bmu}(t, \bfeta) := \frac{\sum_{k\in (\bt^R)\cup (\bt^S)} \mathrm{Var}\left[\widetilde{\bmu}_k(t, \bfeta)\right]^{-1} \widetilde{\bmu}_k(t, \bfeta)}{\sum_{k\in (\bt^R)\cup (\bt^S)} \mathrm{Var}\left[\widetilde{\bmu}_k(t, \bfeta)\right]^{-1}}.
\end{align}

For calculating the weights in Equation \eqref{eq:ODE_estimator}, we need to estimate the unknown variance $\mathrm{Var}[\widetilde{\bmu}_k(t, \bfeta)]$ at a time point $t$ of the ODE solution with initial value $\bmu^{\lbrace R, S \rbrace,k}$. For this, we use all encoded values $\bmu^{\lbrace R, S \rbrace,j}$ for $k < j < t$, i.e., from time points between the current initial time point $k$ and the time point $t$ of interest, and calculate the sample variance $s^2$ of the corresponding ODE solutions $\widetilde{\bmu}_j(t, \bfeta)$ for all $j$ with $k < j < t$.

In our application setting of a clinical registry, typically a more extensive patient characterization at baseline is available, including, e.g., the age at symptom onset or treatment information, which we assume to be informative of intra-individual differences in the underlying disease dynamics.
To obtain personalized dynamics conditional on this baseline information, we use an additional neural network to map the baseline variables to individual ODE parameters $\bfeta_i$, as in our previous work \citep{HacHarPfa2022}. From Equation \eqref{eq:ODE_estimator} we thus obtain estimators of person-specific trajectories $\widetilde{\bmu}_i(t, \bfeta_i)$ for each individual $i$. 

We subsequently evaluate $\widetilde{\bmu}_i(\cdot, \bfeta_i)$ at $(\bt_i^R)$, the individual's measurement time points of measurement instrument $R$, and analogously at $(\bt_i^R)$, the time points of $S$, to obtain the posterior means according to the joint trajectory for both measurement instruments. 
We then sample $\bz_i^{R, k_i} \sim \mathcal{N}(\widetilde{\bmu}_i(k_i, \bfeta_i), \bfsigma_i^{k_i})$ for $k_i \in (\bt^{R}_i)$ for measurement instrument $R$ and pass the sampled latent longitudinal data to the decoder of $\mathrm{VAE}_R$ to obtain reconstructed longitudinal data $\widehat{\bx}_i^R$. Analogously, for $S$, we sample $\bz_i^{S, k_i} \sim \mathcal{N}(\widetilde{\bmu}_i(k_i, \bfeta_i), \bfsigma_i^{k_i})$ for $k_i \in (\bt^S_i)$ and use the decoder of $\mathrm{VAE}_S$ to obtain $\widehat{\bx}_i^S$. The models are trained to provide reconstructed longitudinal data that is as similar as possible to the original observations, to ensure a good fit in the latent space, which allows for a good reconstruction of the original data. %based on which the original data can be reconstructed well. 

\subsection{Optimization based on a joint loss function and adversarial penalty term}
\label{subsec:methods-loss}

We jointly optimize both VAEs and the neural network for obtaining the ODE parameters by maximizing the respective ELBOs of $\mathrm{VAE}_R$ and $\mathrm{VAE}_S$, where we use $\widetilde{\bmu}_i(\bt_i^R, \bfeta_i)$ and $\widetilde{\bmu}_i(\bt_i^S, \bfeta_i)$ as the latent posterior means obtained from the joint ODE solution, respectively. 
Maximizing the ELBOs encourages the model to learn a latent space based on which the original data can be reconstructed well. 
More specifically, an optimal model fit should provide a close alignment of both measurement instruments to the latent ODE while still explaining underlying variance, i.e., it should avoid a trivial solution such as a constant trajectory. 

To enforce a good mapping where the latent representations of each measurement instruments are aligned to the common trajectory,
we propose an additional regularizing penalty term based on the ODE solution. Specifically, we optimize the encoders of each measurement instrument such that it is not possible to distinguish between their latent representations based on their difference to the ODE solution. 
This corresponds to an adversarial domain adaptation approach \citep{Tzeng2017}, where the model is encouraged to find an invariant representation, in which an adversarial classifier is unable to distinguish between domains, i.e., representations of different measurement instruments. Conceptually, this approach is inspired by generative adversarial networks (GANs) \citep{Goodfellow2014}. In a GAN, a generator and a discriminator neural network are trained in a competitive way to learn the distribution of the input data. The generator creates synthetic observations which the discriminator tries to distinguish from real data observations. %The discriminator is trained to optimize this decision, while 
The generator is optimized to fool the discriminator, and thus learns to approximate the data-generating distribution.
In our scenario, we optimize the VAE encoders of both measurement instruments to find a representation in which the latent values from each instrument are not distinguishable based on the ODE solution, i.e., the encoders are trained as generator to fool a discriminator that uses the ODE solution for judging differences between measurement instruments. To achieve this, we compare the respective average differences between each measurement instrument's encoded latent values and the ODE solution across instruments. In particular, we do not use absolute or quadratic deviations, such that encoded values with the same difference above and below the ODE solution cancel out. The ODE solution can thus be viewed as an adversarial classifier that distinguishes between the measurement instruments based on average distance. 
We then add the difference between the per-instrument differences as a penalty term to the loss function, such that the function would be minimized with respect to this term if the difference were zero, i.e., both measurement instruments' encoded values are equally well aligned to the ODE solution. 
To further encourage consistency of each latent representation before and after solving the ODE, we add the squared Euclidean distance between the encoded values $\bmu_i^R$, $\bmu_i^S$ and the posterior means according to the smooth dynamics, $\widetilde{\bmu}_i(\bt_i^R, \bfeta_i)$, $\widetilde{\bmu}_i(\bt_i^S, \bfeta_i)$. 

To ensure that there is no trade-off between a close alignment and explaining variability, and in particular to avoid a trivial constant fit, we encourage similar variances of the ODE solutions and encoded values, separately for $R$ and $S$. Specifically, we add the ratio between the sample variance of all solutions $\widetilde{\bmu}_{i,k}(\bt_i^R, \bfeta_i)$ and the sample variance of all encoded values $\bmu_i^R$, summed across all observed time points, where an offset of $1$ is added to avoid values near zero, and analogously for $S$. 
            
The final loss function is thus given by 
\begin{equation} \label{eq:finalELBO}
	\begin{split}
		\mathcal{L}(\bx_i^R, \bx_i^S, &\bfeta_i, \btheta_R, \bfphi_R, \btheta_S, \bfphi_S) \\
        %&= -\mathrm{ELBO}_{\mathrm{smooth}}(\bx_i, \bfeta_i, \btheta, \bfphi) + \alpha\Vert\bmu_i - \widetilde{\bmu_i}\Vert_2^2 \\
        =&- \mathrm{ELBO}_R(\bx_i^R, \bfeta_i, \btheta_R, \bfphi_R) \\
        &- \mathrm{ELBO}_S(\bx_i^S, \bfeta_i, \btheta_S, \bfphi_S) \\
        &+ \alpha \left( \frac{1}{T_i^R +1} \sum_{k\in (\bt_i^R)} \left( \widetilde{\bmu}_i(k, \bfeta_i) - \bmu_i^{R, k} \right) 
        -  \frac{1}{T_i^S +1} \sum_{k\in (\bt_i^S)} \left( \widetilde{\bmu}_i(k, \bfeta_i) - \bmu_i^{S, k} \right) \right) \\
        &+ \beta \left( \Vert\bmu_i^R - \widetilde{\bmu}_i(\bt_i^R, \bfeta_i)\Vert_2^2 + \Vert\bmu_i^S - \widetilde{\bmu}_i(\bt_i^S, \bfeta_i)\Vert_2^2\right) \\
        &+ \gamma \left( \sum_{k\in (\bt_i^R)} \frac{s^2(\widetilde{\bmu}_i(\bt_i^R, \bfeta_i)) + 1}{s^2(\bmu_i^R) + 1} + \sum_{k\in (\bt_i^S)} \frac{s^2(\widetilde{\bmu}_i(\bt_i^S, \bfeta_i)) + 1}{s^2(\bmu_i^S) + 1} \right) \\
        %&= D_{\mathrm{KL}}(\widetilde{q}_{\bcz \mid \bx_i}(\cdot, \bfeta_i, \bfphi) \Vert p_{\bcz}(\cdot,\btheta)) - \mathrm{E}_{\widetilde{q}}[\log(p_{\bcx \mid \bz_i}(\bx_i,\btheta))] \\
		%&+ \alpha \Vert\bmu_i - \widetilde{\bmu_i}\Vert_2^2 \\
        %&+ \beta \sum_{k=0}^{T_i} \log(s^2((\widetilde{\bmu}_{i,j}(t_k, \bfeta_i))_{j=0,\dots, T_i}) + 0.1) - \log(s^2((\bmu_i^{t_j})_{{j=0,\dots, T_i}}) + 0.1)
	\end{split}
\end{equation}
where $\alpha, \beta, \gamma \in \R_+$ are hyperparameters balancing the loss components.

% modifications: neural net kann auch simple dinge finden, wie konstanten Shift (beruhigend, da kein Overfitting)

% \newpage

\section{Empirical evaluation}
\label{sec:eval}

\subsection{Rare disease registry application}
\label{subsec:eval-data}

To illustrate our approach, we consider longitudinal data on SMA patients' motor function development collected during routine visits into the SMArtCARE registry, a prospective multicenter cohort study \citep{Pechmann2019}. Ethical approval of the study has been granted by the Ethics Committee of the University of Freiburg, No.56/18. So far, the registry provides only few time points per patient with irregular timing and frequency of follow-up visits. 
In addition to an extensive baseline characterization, motoric ability is assessed with physiotherapeutic tests, each comprising several items, at follow-up visits. %, and data on respiratory, nutritional, pain, and adverse events are collected at each visit. 
SMA is caused by a homozygous deletion in the survival motor neuron (SMN) 1 gene on chromosome 5, which is essential for normal motor neuron function \citep{Lefebvre1995}. As a result of the SMN deficiency, muscles do not receive signals from motor neurons, which leads to atrophy, i.e., muscle degeneration. Three different disease-modifying drugs (Nusinersen, Risdiplam, and Onasemnogene Abeparvovec) are available for the treatment of SMA patients that aim to increase production of the SMN protein to maintain functional motor neurons \citep{Schorling2020, Messina2022}. These underlying disease processes cannot be observed directly, but are implicitly reflected in the motor function assessments. %, thus motivating modeling of latent dynamics that drive the observed measurements. 

Specifically, we consider patients treated with Nusinersen who completed the RULM assessment \citep{Mazzone2017} and/or the HFMSE assessment \citep{Glanzman2011}. The HFMSE assessment comprises $33$ items with a maximum score of $66$ to evaluate gross motor function of the whole body and can be conducted for patients with the ability to sit and/or walk older than two years, while the RULM assessment comprises $20$ items with a maximum sum score of $37$ to evaluate motor function of upper limbs and can be conducted for all patients older than two years and able to sit in a wheelchair \citep{Pechmann2019}. 

In our investigation, we use all test items as time-dependent variables and all available baseline information, including, e.g., SMA subtype, age at symptom onset and first treatment, and genetic test results. 
For each patient and each of the two motor function assessments, we removed outlier time points where the difference in HFMSE/RULM sum score to the previous time point exceeded two times the interquartile range of all sum score differences between adjacent time points. For each assessment separately, we filtered out patients with less than two observation time points, and patients with HFMSE/RULM sum score variance smaller than $0.5$, i.e., with nearly constant trajectories across all items. 
This resulted in a dataset of $558$ patients with HFMSE and/or RULM measurements. Of these, $431$ have HFMSE measurements at between $2$ and $15$ time points (median $7$ time points), corresponding to in total $3103$ observations of $33$ time-dependent variables. $427$ patients have RULM measurements at between $2$ and $13$ time points (median $7$ time points), corresponding to in total $2984$ observations of $20$ time-dependent variables. 
In addition, we used $24$ baseline variables. 
Integer-valued test items of both RULM and HFMSE were rescaled and a logit transformation was applied to account for the Gaussian generative distribution parameterized by the VAE decoder.

\subsection{Synthetic dataset scenarios with different error sources}
\label{subsec:eval-modifications}

\begin{figure}[htb]
	\begin{center}
		\includegraphics[width=1\linewidth]{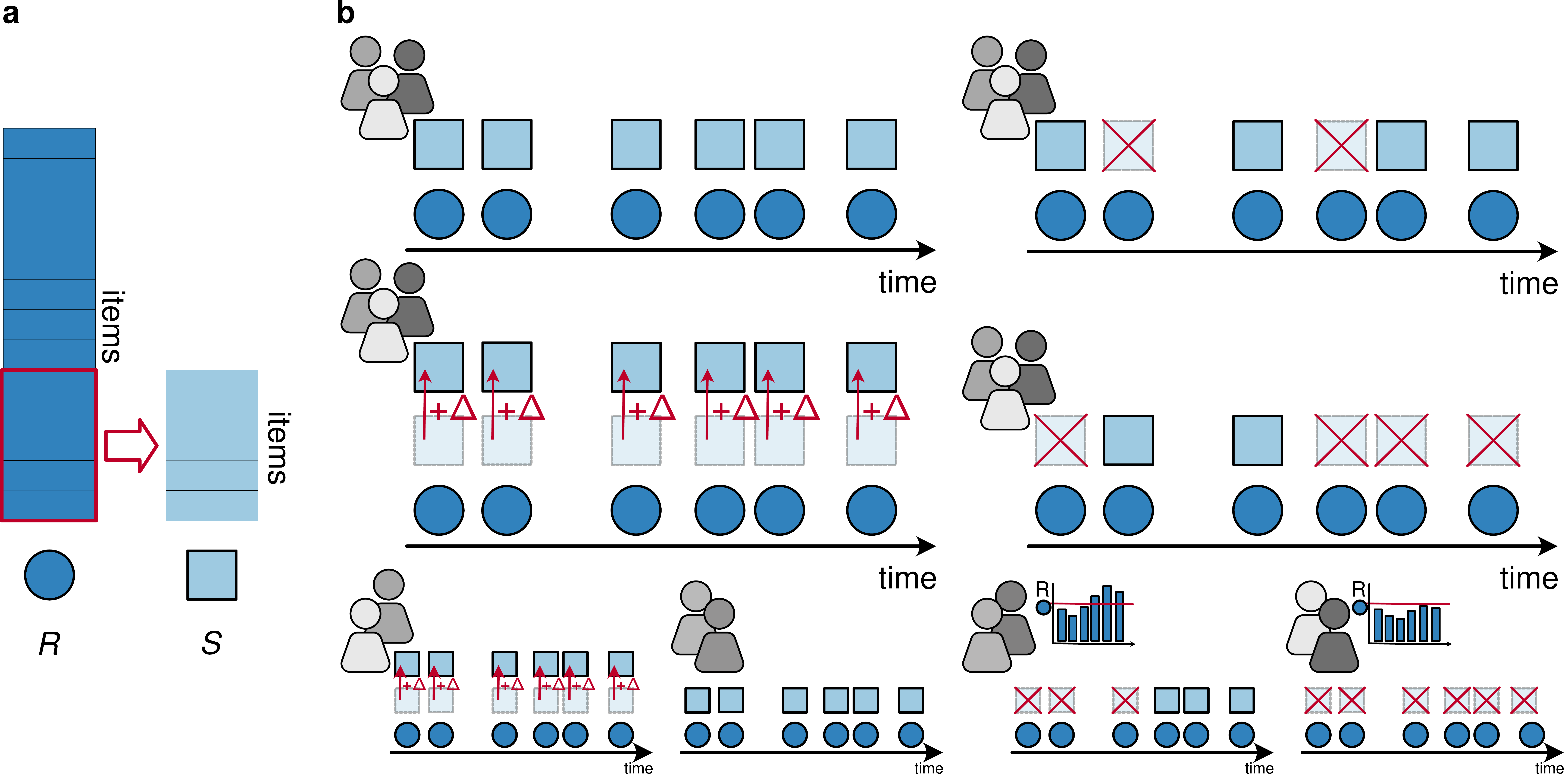}
		\caption{\textbf{a}: Creating an artificial second measurement instrument by using a subscale of a single existing measurement instrument. 
        \textbf{b}: Different dataset modifications: We artificially modify our synthetic second measurement instrument to create different hypothetical scenarios of how discrepancies between instruments could arise. 
        Top left: Baseline scenario with no discrepancies. 
        Top right: Time points are randomly deleted for all patients. 
        Middle left: A shift is added to the items of the second measurement instrument for all patients. 
        Middle right: Time points are deleted preferentially at later time points for all patients.
        Bottom left: A shift is added for a random subgroup of patients. 
        Bottom right: The second measurement instrument is only available if the sum score of the first instrument exceeds a given threshold.}
		\label{fig:modifications}
      \vspace{-0.5cm}
	\end{center}
\end{figure}

To investigate different potential scenarios of systematic differences in measurement instruments and their effect on how well they can be aligned to a common trajectory in a latent space, we create  modifications of the dataset and artificially introduce different sources of discrepancies. 
Specifically, we use data from one measurement instrument only and use a subscale, i.e., a subset of items, as the synthetic second measurement instrument, as schematically illustrated in Figure \ref{fig:modifications}\textbf{a}. 
We use the RULM assessment in the SMArtCARE registry as the original measurement instrument $R$. Thus, from the dataset described in Section \ref{subsec:eval-data}, we consider only the patients with RULM measurements. We use a subscale of the RULM items, comprising $5$ items that concern the handling of small weights as second measurement instrument $S$. 
When no modifications are introduced, the information in our second measurement instrument is fully contained in the first measurement instrument, i.e., a close alignment should be feasible in principle. We consider this scenario as a baseline and then create different modifications of the second measurement instrument, corresponding to the subscale, to reflect different potential sources of systematic discrepancies between measurement instruments. 

The baseline scenario and different modifications are schematically illustrated in Figure \ref{fig:modifications}\textbf{b}. Specifically, we consider two scenarios where a shift is added to all items of the second measurement, reflecting the assumption of a systematic bias in this instrument. 
In the first scenario, the shift is added for all patients and at all time points, reflecting a patient-independent and time-independent source of error (Figure \ref{fig:modifications}\textbf{b}, middle left). 
In the second scenario, the shift is added only for a random subgroup of patients (Figure \ref{fig:modifications}\textbf{b}, bottom left),
%(sampled with $p=0.5$)
reflecting a patient-dependent and time-independent error that could be explained by, e.g., differences in (potentially unobserved) patient characteristics that cause a bias in the measurements. 

Next, we consider scenarios where discrepancies are introduced by 
deleting the information of the second measurement instrument at certain time points. First, we randomly remove observation time points of the second measurement instrument with an equal dropout probability at all time points and for all patients (Figure \ref{fig:modifications}\textbf{b}, top right), to reflect a scenario where the instrument was generally applied less frequently for all patients. 
We then randomly remove observation time points of the second measurement instrument preferentially at later time points, i.e., with a higher dropout probability for later time points, reflecting the assumption of a time-dependent but patient-independent error (Figure \ref{fig:modifications}\textbf{b}, middle right). 
Finally, we consider a scenario where the second measurement instrument is used conditional on a patient's score in the first measurement, i.e., where discrepancies depend on patients' disease development. In particular, we consider observations of the second measurement instrument only if the sum score of all items of the first measurement instrument exceeds a given threshold, and delete all other observations
%remains below a given threshold (we used the $0.75$ quantile of the sum scores of all patients and time points, corresponding to a value of $30$, where $44$ is the maximum sum score), and only include the second measurement at all time points after the sum score exceeded the threshold 
(Figure \ref{fig:modifications}\textbf{b}, bottom right). This reflects the assumption that the second measurement instrument is applied additionally after a ceiling effect is observed in the first measurement instrument, i.e., to still detect changes in a patient's development when the first instrument may not be informative anymore. For example, in the SMArtCARE registry, the RULM test is sometimes used as an additional assessment when ceiling effects in the HFMSE score are observed %, to gain more nuanced information about upper limbs functionality in particular 
\citep[see also][]{Mazzone2017}. 

Implementation details for each modification are provided in the Appendix, Section \ref{subsec:appen-modifications}.

\subsection{Results on synthetically modified datasets}
\label{subsec:eval-modifications-results}

On each of the synthetically created and modified datasets described in the previous Section \ref{subsec:eval-modifications}, we train the model presented in Section \ref{sec:methods}, using separate VAEs for the original RULM measurements as measurement instrument $R$ and the (modified) subscale as second measurement instrument $S$. We use a two-dimensional latent space and a homogeneous linear ODE system, i.e., where $\frac{d}{dt} \bmu(t) = A\cdot \bmu(t) \in \R^2$ and $c=0$. 
We use the same model architecture and hyperparameter configuration for all datasets to ensure a fair comparison (for details, see Section \ref{subsec:appen-implementation}). 
To investigate the effect of the synthetic modifications on the alignment in the latent representation and evaluate the performance of our approach, we consider the misalignment between the latent representations of both measurement instruments after training the model. Specifically, for each patient we calculate the mean absolute difference between the latent representations of each measurement instrument at all time points where both measurement instruments were observed: 

\begin{equation*}
    \Delta_i^{R,S} := \frac{1}{|\bt_i^R \cap \bt_i^S|} \sum_{t \in \bt_i^R \cap \bt_i^S} \mid \bmu_i^{R, t} - \bmu_i^{S,t} \mid \in \R^d.
\end{equation*}

Yet, the absolute values of $\Delta_i^{R,S}$ depend on the scale of the values in the latent representation, which can be flexibly set by the encoder neural network. In addition, what may be considered a large difference depends on the dynamics of the common latent trajectory. For a patient with not much dynamics and a nearly constant ODE trajectory, we would expect smaller differences between the representations of each measurement instrument than for a patient with large changes in dynamics, where the same absolute differences between representations are smaller in relation to the dynamics and might thus be still acceptable. 
To compare the values of $\Delta_i^{R,S}$ across patients, we thus consider them in relation to the variability of each patient's ODE trajectory. Specifically, for each patient we calculate the absolute difference between the values of the ODE solution at the first and last time point of the time interval, 

\begin{equation*}
    \Delta_i^{\mathrm{ODE}} := \mid \widetilde{\bmu}_i(t_{\max\lbrace T_i^R, T_i^S \rbrace}), \bfeta_i) - \widetilde{\bmu}_i(t_0, \bfeta_i) \mid \in \R^d.
\end{equation*}

For each scenario, we calculate $\Delta_i^{R,S}$ and $\Delta_i^{\mathrm{ODE}}$ for all patients $i=1, \dots, N$ and visualize them in scatterplots. Ideally, the variability in ODE dynamics should be greater than the discrepancies between measurement instruments, i.e., the better the alignment, the more points lie above the diagonal. 

First, we focus on the baseline scenario where no modification was introduced in the subscale. This scenario can be viewed as a performance reference for the following scenarios with modifications, as all the information of the second measurement instrument is already contained in the first one, such that a close alignment should be feasible in principle. 
Nonetheless, due to the random initialization of each model and the iterative optimization of a non-convex loss function, empirically it might not be possible to achieve a perfect mapping. 
This scenario thus allows to quantify potential discrepancies due to other reasons. 

\begin{figure}[htb]
	\begin{center}
		\includegraphics[width=1\linewidth]{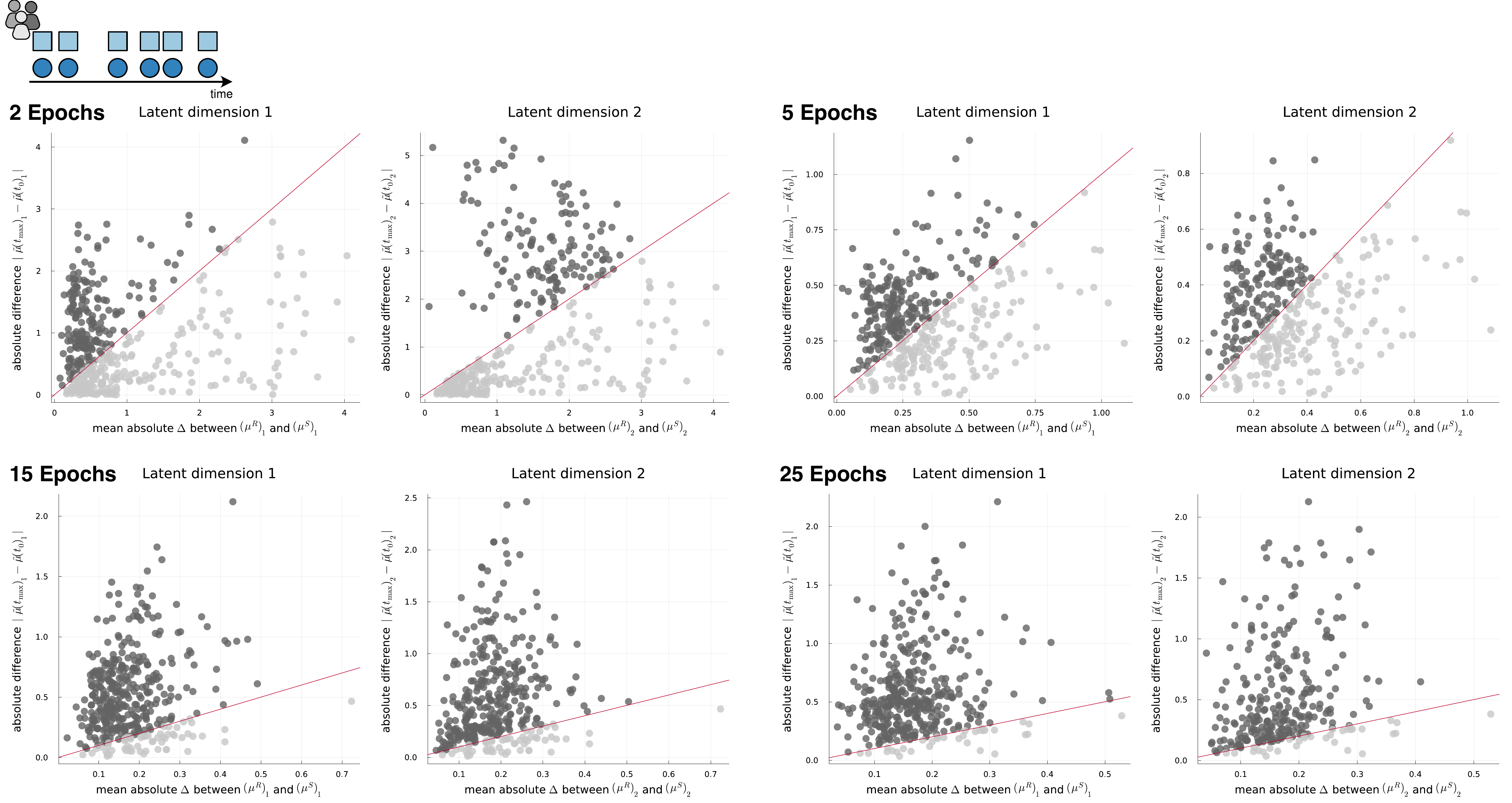}
		\caption{Evolution of the alignment of measurement instruments in the baseline scenario where no modification was applied over the course of $25$ training epochs. 
        Each subplot corresponds to the state of the model after training for the specified number of epochs. The two panels of each subplot correspond to the two latent dimensions. 
        In each panel, we show the values of $(\Delta_i^{R,S})_i$, i.e., the mean absolute difference between the latent representations of the two measurement instruments in the first resp. second latent dimension, on the $x$ axis and the corresponding values of $(\Delta_i^{\mathrm{ODE}})_i$, i.e., the differences in ODE dynamics in the first resp. second latent dimension, on the $y$ axis. 
        The solid red line corresponds to the diagonal, and the points above the diagonal, i.e., where $\Delta_i^{\mathrm{ODE}} > \Delta_i^{R,S}$ are colored in dark grey, while the points below the diagonal where $\Delta_i^{\mathrm{ODE}} < \Delta_i^{R,S}$ are colored in light grey.
        }
		\label{fig:no-modification-results}
      \vspace{-0.5cm}
	\end{center}
\end{figure} 

In Figure \ref{fig:no-modification-results}, we visualize the changes in the $(\Delta_i^{R,S}, \Delta_i^{\mathrm{ODE}})_{i=1, \dots, N}$ values over the course of the first $25$ training epochs, separately for each of the two latent dimensions. This might also be used as a reference for assessing performance later in more complex scenarios. At the beginning of training, for many patients the difference between the two measurement instruments is greater than the variability explained by the ODE dynamics (indicated by many values below the diagonal). As training progresses, this ratio changes (indicated by more and more values below the diagonal), until after $25$ epochs, for nearly all patients the difference between the measurement instruments is (often considerably) smaller than the changes in ODE dynamics. 
Note that also the scale of the differences between measurement instruments changes from $\left[ 0, 4 \right]$ to $\left[ 0, 0.5 \right]$, i.e., not only the ratio of the two values changes, but also the absolute values of $(\Delta_i^{R,S})_i$ become much smaller. Thus, in a scenario where a close alignment is feasible, the model is effective in minimizing the differences between measurement instruments and achieves a very close, yet not perfect mapping.

\begin{figure}[H]
	\begin{center}
		\includegraphics[width=1\linewidth]{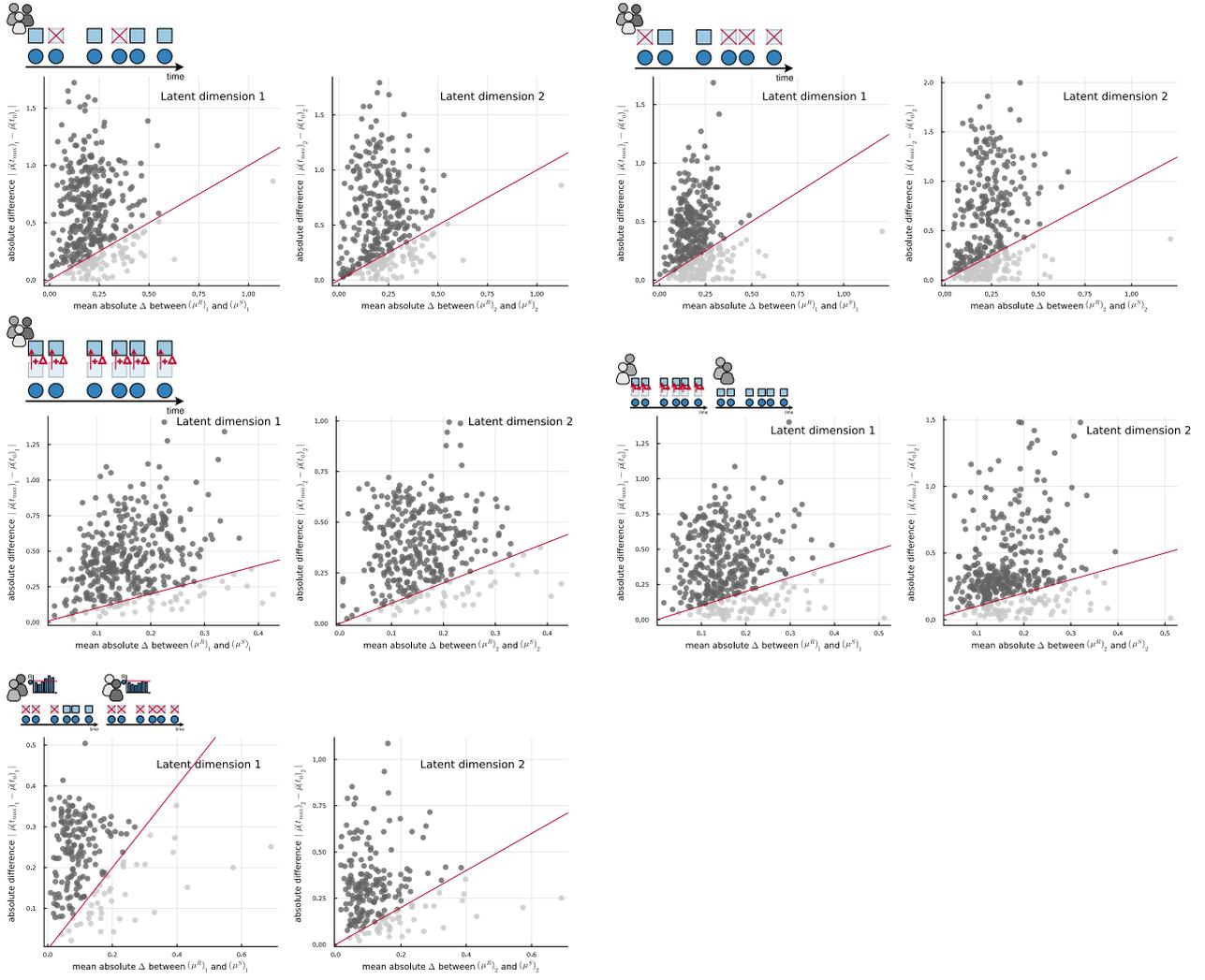}
		\caption{Results of model training on different dataset modifications with artificially introduced discrepancies. Each of the five subplots corresponds to a specific dataset modification, symbolized by the icon in the top left corner (see Figure \ref{fig:modifications} and the corresponding explanation in Section \ref{subsec:eval-modifications}). The two panels of each subplot correspond to the two dimensions of the latent representation.  
        In each panel, we show the values of $(\Delta_i^{R,S})_i$, i.e., the mean absolute difference between the latent representations of the two measurement instruments in the first resp. second latent dimension, on the $x$ axis and the corresponding values of $(\Delta_i^{\mathrm{ODE}})_i$, i.e., the differences in ODE dynamics in the first resp. second latent dimension, on the $y$ axis. 
        The solid red line corresponds to the diagonal, and the points above the diagonal, i.e., where $\Delta_i^{\mathrm{ODE}} > \Delta_i^{R,S}$ are colored in dark grey, while the points below the diagonal where $\Delta_i^{\mathrm{ODE}} < \Delta_i^{R,S}$ are colored in light grey.}
		\label{fig:modifications-results}
      \vspace{-0.5cm}
	\end{center}
\end{figure}

Figure \ref{fig:modifications-results} shows scatterplots of each patient's $\Delta_i^{R,S}$ and $\Delta_i^{\mathrm{ODE}}$ values after model training in all scenarios with a synthetic modification of the second measurement instrument described in Section \ref{subsec:eval-modifications}. 

In the first row, we show results for two modifications where for all patients, time points of the second measurement instrument were deleted at random time points (left) or preferentially at later time points (right). In both scenarios, for most patients $\Delta_i^{R,S}$ is smaller than $\Delta_i^{\mathrm{ODE}}$, and the absolute values of $(\Delta_i^{R,S})_i$ are mostly in a similar range as in the baseline scenario (almost all values below $0.5$, two/four values between $0.5$ and $0.7$ and one outlier $> 1$). 
Deleting items with a higher probability at later time points corresponds to a more challenging task, as the model has to deal with potentially longer periods of missing information, and has to extrapolate rather than interpolate. Yet, the number of values above the diagonal is only slightly higher in this scenario, indicating that the approach is generally capable of handling also these more demanding challenges.  

In the second row, we show results for applying a shift to all items of the second measurement instrument at all time points either for all patients (left) or for a randomly sampled subgroup of patients (right). In both scenarios, the number of points below the diagonal is smaller than for the scenarios where time points were deleted, and in a similar range as in the baseline scenario (all values $< 0.5$, only one outlier in the second scenario slightly larger than $0.5$). Particularly in the first scenario of a patient-independent shift, the alignment is only slightly worse than in the baseline scenario with no modification. This shows that a simple additive transformation can be easily handled by the model, even if it is only present in a subgroup. %TODO: discuss no overfitting?!

In the third row, we show results for the scenario where the observation of the second measurement depends on the sum score of all items of the first measurement instrument. %In the left panel, we consider a scenario where observation time points of the second measurement instrument are deleted if the sum score of the first is above a given threshold, while in the right panel we consider the opposite scenario where the second measurement instrument is only observed if the sum score of the first exceeds the threshold. 
Specifically, the second measurement instrument is only observed if the sum score of the first measurement instrument exceeds the threshold, i.e., is applied when ceiling effects are detected in the first measurement instrument. 
%In the first scenario, still a close mapping is achieved for most patients, similar to or even slightly better than in the scenario where time points are preferentially deleted at later time points for all patients. The second scenario is considerably more challenging, 
This scenario is considerably more challenging, as the second measurement instrument is missing completely for a subgroup of patients (as can be seen by the overall smaller number of points in the scatterplot). The scale of the $(\Delta_i^{R,S})_i$ values is still comparable to the other scenarios, with only $2$ values $>0.5$. Yet, here the scale of the $(\Delta_i^{\mathrm{ODE}})_i$ values is smaller, in particular in the first dimension, i.e., the ODE solution is more strongly regularized, as overall less information is available for fitting. Still, also in this challenging scenario, for most patients $\Delta_i^{R,S} < \Delta_i^{\mathrm{ODE}}$, i.e., a decent alignment is still feasible when the second measurement instrument is observed dependent on a patient's status, and generally much less frequently.

\subsection{Results on two different measurement instruments in the SMArtCARE registry}
\label{subsec:eval-SMArtCARE}

We apply our model in a real data example with two real measurement instruments, i.e., without artificial modifications, where discrepancies between measurement instruments could arise due to more complex reasons. Specifically, we consider data from the HFMSE and RULM measurement instruments in the SMArtCARE data as described in Section \ref{subsec:eval-data}. Measurement instrument $R$ corresponds to patients' HFMSE measurements, and measurement instrument $S$ corresponds to patients' RULM measurements. 

On this dataset, we compare different versions of penalties in our approach with respect to their effectiveness in achieving a close alignment. Correspondingly, we train the model with four different loss function versions. We use (1) neither the adversarial penalty term nor the penalty for aligning each measurement's representation separately to the ODE trajectory (referred to as the 'ODE penalty' in the following), i.e., $\alpha=\beta=0$. Then, we train models with (2) only the ODE penalty, i.e., $\alpha=0, \beta \neq 0$, and (3) only the adversarial penalty, i.e., $\alpha \neq 0, \beta=0$, and finally (4) with both penalty terms, i.e., $\alpha, \beta \neq 0$. All other hyperparameters are kept constant across the four types of model estimation to ensure comparability. 

\begin{figure}[htb]
	\begin{center}
		\includegraphics[width=1\linewidth]{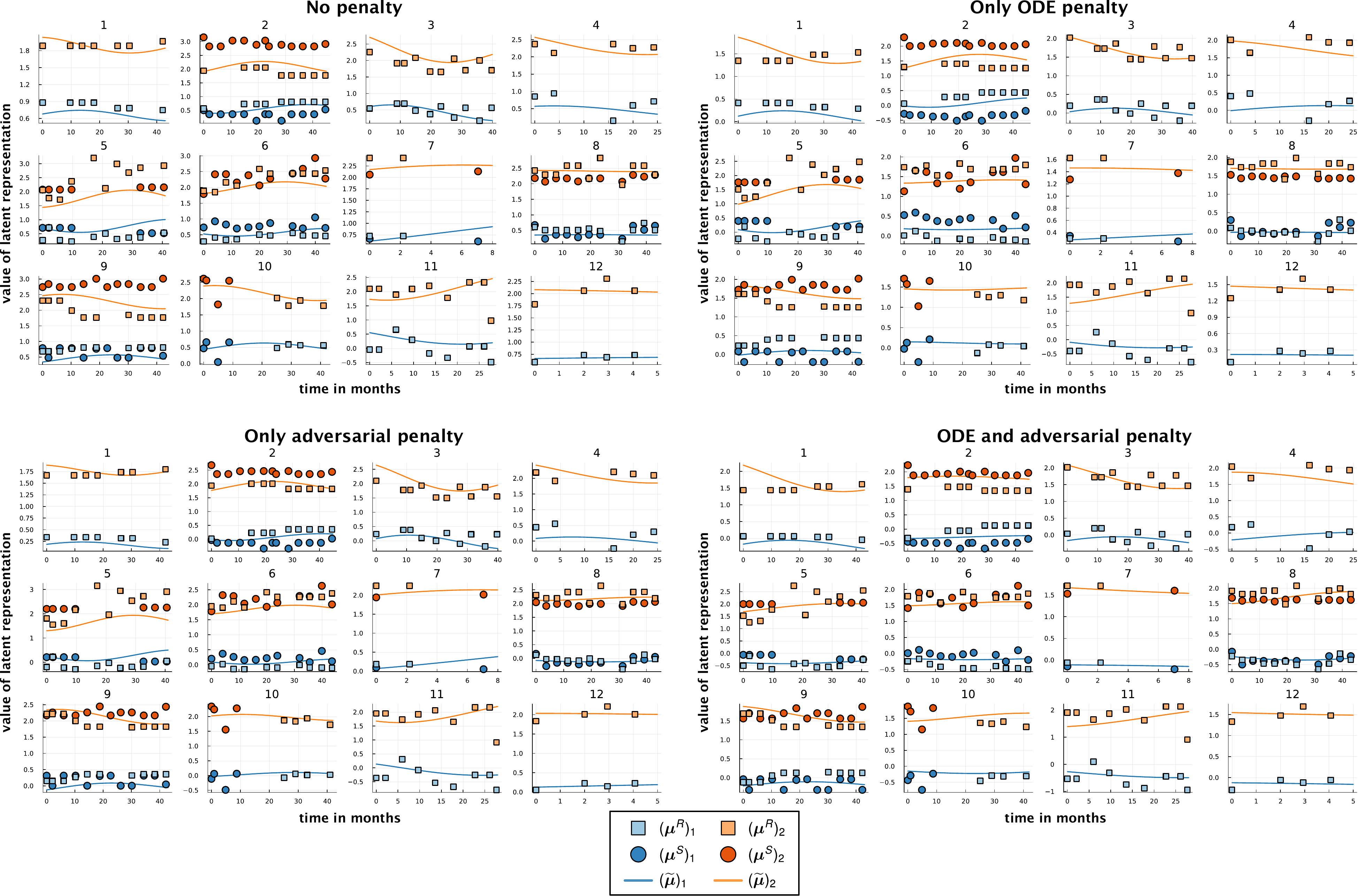}
		\caption{Exemplary model fits of $12$ randomly selected individuals from the SMArtCARE registry for four model fitting versions, using different combinations of penalty terms for alignment, shown in subplots. In each subplot, each panel shows the latent representation and fitted ODE trajectory of the data from one patient, the $x$ axis corresponds to the time in months and the $y$ axis to the value of the latent representation. The $2$-dimensional latent representations $\bmu_i^R$ obtained directly from the encoder of the HFMSE measurements are depicted as squares in light colors, blue for the first dimension and orange for the second dimension, whereas the latent representations $\bmu_i^S$ obtained from the RULM encoder are depicted as circles in darker colors, again using blue for the first dimension and orange for the second dimension  The common ODE trajectory $\widetilde{\bmu}_i$ is shown as blue resp. orange line for the first resp. second latent dimension. 
		}
		\label{fig:penalties-results}
      \vspace{-0.5cm}
	\end{center}
\end{figure}

In Figure \ref{fig:penalties-results}, we visualize exemplary model fits for $12$ randomly selected individuals from the SMArtCARE registry obtained for the four considered penalty approaches. Applying the ODE penalty only (Figure \ref{fig:penalties-results}, top right) helps to reduce the distances between the latent representations $\bmu_i^R$, $\bmu_i^S$ to the common ODE trajectory $\widetilde{\bmu}(t, \bfeta)$. The adversarial penalty is even more effective in achieving a close alignment between the representations of the two different measurement instruments (Figure \ref{fig:penalties-results}, bottom left). When both terms are applied together, the closest alignment is achieved. Yet, the ODE penalty also has a regularizing effect on the fitted dynamics, i.e., the ODE trajectories tend to become closer to constant if that term is added, whereas the adversarial penalty term has no effect on the variance explained by the ODE solution. In this specific setting, we would thus recommend to only use the adversarial penalty term for obtaining a representation that provides a good alignment yet still explains underlying variance. 

\clearpage

\section{Discussion}
\label{sec:discussion}

%TODO: say more explicitly: we assume a common underlying process, that is captured by the ODE? 

Domain adaptation is a prominent concept in computer science, but has rarely been considered in statistical modeling of longitudinal data so far. In particular, it is implicitly assumed that an underlying latent concept is shared between data from two different domains, and a fully domain-invariant mapping is in principle feasible. Yet, in more complex clinical cohort applications such as a rare disease registry with small numbers of patients and few and irregular time points per patient, it is not clear to what extent different measurement instruments can be aligned or whether no useful mapping is possible. 

To investigate the potential of domain adaptation for such settings, we have therefore developed and evaluated an approach that is tailored for mapping longitudinal measurements from different measurement instruments in a clinical registry with few time points. 
As a motivating application, we have considered different measurement instruments used to assess motor function of individuals in an SMA registry at different time points. 
For aligning the different motor function measurement instruments, each comprising a larger number of individual items, we have separately mapped the items of each instrument to a latent space using a VAE, for learning a flexible low-dimensional representation based on neural networks. 
We have allowed for individual trajectories by inferring person-specific ODE parameters from baseline variables with an additional neural network, jointly optimizing the VAEs for the different measurement instruments, the neural network for inferring ODE parameters, and the dynamic model in the latent space. This is to find a representation that is optimized to provide a good mapping between measurement instruments. 
To enforce a close alignment during model optimization, we have developed an additional penalty term, inspired by adversarial classifiers in domain adaptation and GANs, which encourages the model to find a latent space where the representations of different measurement instruments cannot be distinguished based on the ODE solution. 
To specifically investigate what happens in scenarios when no perfect mapping can be achieved, we have designed synthetic modifications of an SMA dataset by artificially introducing discrepancies with varying degrees of complexity, to characterize the effect of such systematic discrepancies on the alignment. 

Our findings show that our proposal, as a prototypical domain adaptation approach, can achieve a close alignment in simple settings, and still provides a reasonable mapping in complex settings, e.g., when one measurement instrument is observed only if the sum score of another measurement instrument exceeds a given threshold, i.e., when it is observed dependent on a patient's status and not at all for many patients. 
To exemplify how domain adaptation approaches could be adapted in a real world setting, where discrepancies might arise due to even more intricate reasons, we have considered data from two different measurement instruments in the SMArtCARE registry. We investigated different types of model estimation in our approach, using different penalty terms. There, we could show that the adversarial penalty was particularly important for achieving a close alignment, and that a reasonable, yet not perfect, mapping was achievable despite the increased complexity. This application also illustrates potential trade-offs in complex settings where a perfect mapping might not be available, as enforcing a closer alignment to the ODE solution with an ODE penalty term also implied more strongly regularized ODE solutions, and thus less explained variance in the dynamics. 

%In a setting where there is only little overlap between the information of different measurement instruments, instead of accepting a certain amount of discrepancy in the mapping, it might be more desirable to explicitly model shared and distinct components separately. This requires to decide whether or to what extent a mapping is feasible, which can be addressed, e.g., by investigating synthetic modifications or model versions as we have illustrated.

In our application, we have still observed differences in the alignment across individuals, and also between different observed time intervals for the same individual, suggesting that there are time-dependent and time-independent sources of intra-individual differences which prevent a closer alignment. 
While in the present manuscript, we have focused on investigating the general potential of a domain adaptation approach, these differences might need to be characterized in more detail in future work. For example, the observed patient-and time-dependent differences between measurement instruments could be linked to patient characteristics observed in the baseline variables. 
Further, the choice of the ODE system used for obtaining the common trajectory might influence the quality of the alignment, e.g., depending on whether an interaction between different latent dimensions is possible or not. Due to the small number of patients, we have focused on a simple linear system and have not investigated the effect of different choices of ODE systems in depth so far. 

In summary, we have illustrated how a domain adaptation approach can be adapted to address the challenge of different measurement instruments in clinical cohorts with few time points, and how the performance of such an approach can be characterized. Specifically, combining deep learning components and dynamic modeling by differential equations allowed for aligning the representations of measurement instruments in a latent space, indicating that such a combination of approaches could be more generally useful for addressing challenging modeling scenarios in longitudinal data, such as from rare disease registries. Overall, our results suggest that domain adaptation techniques might have considerable potential for more generally integrating information of different measurement instruments.

% This nuanced exploration is crucial for understanding the limitations and boundaries of the proposed domain adaptation approach.

\vspace{\baselineskip}

\noindent {\bf{Acknowledgement}}
This work was funded by the Deutsche Forschungsgemeinschaft (DFG, German Research Foundation) -- Project-ID 322977937 -- GRK 2344 (MH) and Project-ID 499552394 -- SFB 1597 (HB, MB, and MH). Biogen and Novartis provide financial support for the SMArtCARE registry. AP was supported by the Berta-Ottenstein clinician scientist program of the University of Freiburg.

\vspace{\baselineskip}

\noindent {\bf{Conflict of Interest}}
	
\noindent {\it{The authors have declared no conflict of interest.}}

\renewcommand\thesubsection{A.\arabic{subsection}}    

\section*{Appendix}

\setcounter{section}{0}    

\subsection{Implementation details}
\label{subsec:appen-implementation}

The following section provides details on the implementation of the proposed domain adaptation approach.

The code to run all experiments and models is written in the Julia programming language \citep{Bezanson2017} of version 1.8.3 with the additional packages {\tt{CSV.jl}} (v0.10.11), {\tt{DataFrames.jl}} (v1.6.1), {\tt{Distances.jl}} (v0.10.10), {\tt{Distributions.jl}} (v0.25.102), {\tt{Flux.jl}} (v0.13.17), {\tt{LaTeXStrings.jl}} (v1.3.0), {\tt{Measures}} (v.0.3.2), {\tt{MultivariateStats}} (v.0.10.2), {\tt{Optim.jl}} (v1.7.8),
{\tt{OrderedCollections.jl}} (v1.6.2), {\tt{Parameters.jl}} (v0.12.3), {\tt{Plots.jl}} (v1.23.5), {\tt{ProgressMeter.jl}} (v1.39.0), {\tt{StatsBase.jl}} (v0.34.2), {\tt{VegaLite.jl}} (vv3.2.3), and {\tt{Zygote.jl}} (v.0.6.59). 

In the VAEs for each measurement instrument, the encoder and decoder each have one hidden layer, where the number of hidden units is equal to the number of input dimensions, i.e., the number of variables in each dataset, corresponding to the number of items in each measurement instrument. 
In all hidden layers, we use a $\tanh$-activation function. 
Each latent space is two-dimensional, and the respective mean and variance are obtained as affine linear transformations of the hidden layer values without a non-linear activation. 
Each decoder parameterizes mean and variance of a Gaussian distribution, corresponding to the conditional distribution of the input data given the latent representation, calculated from the decoder hidden layer using an affine linear transformation. 
	
The additional network for mapping the baseline variables to individual-specific ODE parameters has two hidden layers in addition to input and output layer. In the first hidden layer, the number of hidden units is equal to the number of baseline variables and a $\tanh$-activation is used. In the second hidden layer, the number of hidden units is equal to the number of ODE parameters and the activation function is a sigmoid function, shifted by $-0.5$ and scaled by $0.5$. This corresponds to a prior for the range of the estimated ODE parameters. Deviations from this range are still possible, as an affine linear transformation with a diagonal matrix is added as the final layer. 

In the ELBO of each VAE, the KL-divergence between the prior and posterior is scaled by a factor of $0.5$ to slightly reduce the regularizing effect of the prior. The sum of the squared decoder parameter is added with a weighting factor of $0.01$ to the loss function, corresponding to a standard penalty term to prevent exploding decoder parameters. 

In the experiments on synthetic modifications of data from one measurement instrument presented in Section \ref{subsec:eval-modifications-results}, we use a two-dimensional latent space and a homogeneous linear ODE system, i.e., where $\frac{d}{dt} \bmu(t) = A\cdot \bmu(t)\in \R^2$ and $c=0$. 
All additional loss function penalty terms are added with a weight of $5$, i.e., in the notation of Equation \eqref{eq:finalELBO}, we have $\alpha = \beta = \gamma = 5$. 
In the application on data from two measurement instruments in the SMArtCARE registry presented in Section \ref{subsec:eval-SMArtCARE}, we also use a two-dimensional latent space and an inhomogeneous linear ODE system, i.e., where $\frac{d}{dt} \bmu(t) = A\cdot \bmu(t) + c \in \R^2$, to account for the higher complexity of this setting. In the loss function, we set $\gamma = 5$ in all four types of model estimation. In the model with neither ODE penalty nor adversarial penalty, we set $\alpha = \beta = 0$. In the model with only the ODE penalty, we set $\alpha = 0, \beta = 5$, while in the model with only the adversarial penalty, we set $\alpha = 5, \beta = 0$. In the model with both ODE and adversarial penalty, we set $\alpha = \beta = 5$
 
All models are trained using on stochastic gradient descent with the ADAM optimizer \citep{Kingma2015}. 
For the experiments on the dataset modifications from Section \ref{subsec:eval-modifications-results}, we use a learning rate of $0.001$ and $30$ training epochs for all modifications, chosen based on monitoring convergence of the loss function and visualizing exemplary fits. 
For the results on two measurement instruments in the SMArtCARE registry from Section \ref{subsec:eval-SMArtCARE}, we use a learning rate of $0.00003$ and $10$ training epochs for all four versions of model estimation.  

\subsection{Dataset modifications}
\label{subsec:appen-modifications}

In the following, we provide details of the implementation of each modification presented in Section \ref{subsec:eval-modifications}. 
For the scenarios with a shift, we add an offset of $2$ to all items. 
In the scenario where the shift is only applied to a random subgroup of patients, we sample the subgroup with $p=0.5$. 
In the scenario where observation time points of the second measurement instrument are deleted randomly for all patients, we use a dropout probability of $0.5$. 
For removing observation time points preferentially at later time points, for a patient $i$, at each time point $k = 1, \dots, T_i^R+1$ we sample from a Bernoulli distribution with $p = \frac{k}{T_i^R+4}$ and remove the time points where a $1$ was sampled, such that removal of later time points is more likely. 
For removing observations in the second measurement instrument where the sum score of all items of the first measurement exceeds a given threshold, we use as cutoff the $0.6$ quantile of the sum scores of all patients and time points, corresponding to a value of $25$, where $44$ is the maximum sum score.  

\small{
\bibliographystyle{agsm}
\bibliography{bibfile.bib}

@article{Blei2017,
	author = {David M. Blei and Alp Kucukelbir and Jon D. McAuliffe},
	title = {Variational Inference: A Review for Statisticians},
	journal = {Journal of the American Statistical Association},
	volume = {112},
	number = {518},
	pages = {859--877},
	year  = {2017},
	publisher = {Taylor & Francis}
}

@InProceedings{Kingma2014,
	author    = {Diederik P. Kingma and Max Welling},
	editor    = {Yoshua Bengio and Yann LeCun},
	title     = {Auto-Encoding Variational {B}ayes},
	booktitle = {2nd International Conference on Learning Representations ({ICLR}),
	Conference Track Proceedings},
	year      = {2014},
}

@article{Kingma2019,
	title={An Introduction to Variational Autoencoders},
	volume={12},
	ISSN={1935-8245},
	number={4},
	journal={Foundations and Trends® in Machine Learning},
	publisher={Now Publishers},
	author={Kingma, Diederik P. and Welling, Max},
	year={2019},
	pages={307--392}
}

@InProceedings{Kingma2015,
	author = {Diederik P. Kingma and Jimmy Ba},
	editor    = {Yoshua Bengio and Yann LeCun},
	title     = {Adam: {A} Method for Stochastic Optimization},
	booktitle = {3rd International Conference on Learning Representations ({ICLR}),
	Conference Track Proceedings},
	year      = {2015},
}

@article{Bezanson2017, 
	title = {Julia: A Fresh Approach to Numerical Computing},
	author={Jeff Bezanson and Alan Edelman and Stefan Karpinski and Viral B. Shah},
	journal = {SIAM Review},
	year = {2017},
	volume = {59},
	number = {1},
	pages = {65--98}
}

@article{HacHarPfa2022,
    Author={Hackenberg, M.  and Harms, P.  and Pfaffenlehner, M.  and Pechmann, A.  and Kirschner, J.  and Schmidt, T.  and Binder, H. },
    Title={{D}eep dynamic modeling with just two time points: Can we still allow for individual trajectories?},
    Journal={Biometrical Journal},
    volume = {64},
    pages = {1426–1445}, 
    doi = {https://doi.org/10.1002/bimj.202000366},
    Year={2022},
}

@misc{Hackenberg2023,
	title = {A statistical approach to latent dynamic modeling with differential equations},
    note={arXiv preprint: https://arxiv.org/abs/2311.16286},
	doi = {10.48550/arXiv.2311.16286},
	publisher = {arXiv},
	author = {Hackenberg, Maren and Pechmann, Astrid and Kreutz, Clemens and Kirschner, Janbernd and Binder, Harald},
	year = {2023},
}

@article{Shahar2017, 
    author = {Doron J. Shahar},
    title = {Minimizing the Variance of a Weighted Average},
    year = {2017},
    journal = {Open Journal of Statistics},
    volume = {7},
    number = {2},
    pages = {216-224},
    doi = {10.4236/ojs.2017.72017}
}

@article{Guan2022,
  author={Guan, Hao and Liu, Mingxia},
  journal={IEEE Transactions on Biomedical Engineering}, 
  title={Domain Adaptation for Medical Image Analysis: A Survey}, 
  year={2022},
  volume={69},
  number={3},
  pages={1173-1185},
  doi={10.1109/TBME.2021.3117407}
}

@misc{Csurka2017,
	title = {Domain {Adaptation} for {Visual} {Applications}: {A} {Comprehensive} {Survey}},
	shorttitle = {Domain {Adaptation} for {Visual} {Applications}},
    eprint = {1702.05374}, 
    archivePrefix = {arXiv},
    note = {arXiv preprint: https://arxiv.org/abs/1702.05374},
	doi = {10.48550/arXiv.1702.05374},
	publisher = {arXiv},
	author = {Csurka, Gabriela},
	year = {2017},
}

@inproceedings{Farahani2021,
	address = {Cham},
	series = {Transactions on {Computational} {Science} and {Computational} {Intelligence}},
	title = {A {Brief} {Review} of {Domain} {Adaptation}},
	doi = {10.1007/978-3-030-71704-9_65},
	booktitle = {Advances in {Data} {Science} and {Information} {Engineering}},
	publisher = {Springer International Publishing},
	author = {Farahani, Abolfazl and Voghoei, Sahar and Rasheed, Khaled and Arabnia, Hamid R.},
	editor = {Stahlbock, Robert and Weiss, Gary M. and Abou-Nasr, Mahmoud and Yang, Cheng-Ying and Arabnia, Hamid R. and Deligiannidis, Leonidas},
	year = {2021},
	pages = {877--894},
}

@inproceedings{Karani2018,
	address = {Cham},
	series = {Lecture {Notes} in {Computer} {Science}},
	title = {A {Lifelong} {Learning} {Approach} to {Brain} {MR} {Segmentation} {Across} {Scanners} and {Protocols}},
	doi = {10.1007/978-3-030-00928-1_54},
	booktitle = {Medical {Image} {Computing} and {Computer} {Assisted} {Intervention} – {MICCAI} 2018},
	publisher = {Springer International Publishing},
	author = {Karani, Neerav and Chaitanya, Krishna and Baumgartner, Christian and Konukoglu, Ender},
	editor = {Frangi, Alejandro F. and Schnabel, Julia A. and Davatzikos, Christos and Alberola-López, Carlos and Fichtinger, Gabor},
	year = {2018},
	pages = {476--484},
}

@inproceedings{Ghafoorian2017,
	address = {Cham},
	series = {Lecture {Notes} in {Computer} {Science}},
	title = {Transfer {Learning} for {Domain} {Adaptation} in {MRI}: {Application} in {Brain} {Lesion} {Segmentation}},
	shorttitle = {Transfer {Learning} for {Domain} {Adaptation} in {MRI}},
	doi = {10.1007/978-3-319-66179-7_59},
	booktitle = {Medical {Image} {Computing} and {Computer} {Assisted} {Intervention} - {MICCAI} 2017},
	publisher = {Springer International Publishing},
	author = {Ghafoorian, Mohsen and Mehrtash, Alireza and Kapur, Tina and Karssemeijer, Nico and Marchiori, Elena and Pesteie, Mehran and Guttmann, Charles R. G. and de Leeuw, Frank-Erik and Tempany, Clare M. and van Ginneken, Bram and Fedorov, Andriy and Abolmaesumi, Purang and Platel, Bram and Wells, William M.},
	editor = {Descoteaux, Maxime and Maier-Hein, Lena and Franz, Alfred and Jannin, Pierre and Collins, D. Louis and Duchesne, Simon},
	year = {2017},
	pages = {516--524},
}

@article{Goetz2016,
	title = {{DALSA}: {Domain} {Adaptation} for {Supervised} {Learning} {From} {Sparsely} {Annotated} {MR} {Images}},
	volume = {35},
	shorttitle = {{DALSA}},
	doi = {10.1109/TMI.2015.2463078},
	number = {1},
	journal = {IEEE Transactions on Medical Imaging},
	author = {Goetz, Michael and Weber, Christian and Binczyk, Franciszek and Polanska, Joanna and Tarnawski, Rafal and Bobek-Billewicz, Barbara and Koethe, Ullrich and Kleesiek, Jens and Stieltjes, Bram and Maier-Hein, Klaus H.},
	year = {2016},
	pages = {184--196},
}

@article{Becker2015,
	title = {Domain {Adaptation} for {Microscopy} {Imaging}},
	volume = {34},
    doi = {10.1109/TMI.2014.2376872},
	number = {5},
	journal = {IEEE Transactions on Medical Imaging},
	author = {Becker, Carlos and Christoudias, C. Mario and Fua, Pascal},
	year = {2015},
	pages = {1125--1139},
}

@inproceedings{Tzeng2017,
	address = {Honolulu, HI},
	title = {Adversarial {Discriminative} {Domain} {Adaptation}},
	doi = {10.1109/CVPR.2017.316},
	booktitle = {2017 {IEEE} {Conference} on {Computer} {Vision} and {Pattern} {Recognition} ({CVPR})},
	publisher = {IEEE},
	author = {Tzeng, Eric and Hoffman, Judy and Saenko, Kate and Darrell, Trevor},
	year = {2017},
	pages = {2962--2971},
}

@article{Chen2012,
	title = {Marginalized {Denoising} {Autoencoders} for {Domain} {Adaptation}},
    note = {arXiv preprint: https://arxiv.org/abs/1206.4683},
	journal = {ArXiv},
	author = {Chen, Minmin and Xu, Z. and Weinberger, Kilian Q. and Sha, Fei},
	year = {2012},
}

@article{Long2015,
	title = {Learning {Transferable} {Features} with {Deep} {Adaptation} {Networks}},
    note = {arXiv preprint: https://arxiv.org/abs/1502.02791},
	journal = {ArXiv},
	author = {Long, Mingsheng and Cao, Yue and Wang, Jianmin and Jordan, Michael I.},
	year = {2015},
}

@inproceedings{Ganin2015,
	title = {Unsupervised {Domain} {Adaptation} by {Backpropagation}},
	booktitle = {Proceedings of the 32nd {International} {Conference} on {Machine} {Learning}},
	publisher = {PMLR},
	author = {Ganin, Yaroslav and Lempitsky, Victor},
	year = {2015},
	pages = {1180--1189},
}

@inproceedings{Purushotham2016,
  title={Variational Recurrent Adversarial Deep Domain Adaptation},
  author={S. Purushotham and Wilka Carvalho and Tanachat Nilanon and Yan Liu},
  booktitle={International Conference on Learning Representations},
  year={2016},
}

@misc{Jude2022,
	title = {Robust alignment of cross-session recordings of neural population activity by behaviour via unsupervised domain adaptation},
    note = {arXiv preprint: http://arxiv.org/abs/2202.06159},
	doi = {10.48550/arXiv.2202.06159},
	publisher = {arXiv},
	author = {Jude, Justin and Perich, Matthew G. and Miller, Lee E. and Hennig, Matthias H.},
	year = {2022},
}

@misc{He2023,
	title = {Domain {Adaptation} for {Time} {Series} {Under} {Feature} and {Label} {Shifts}},
    note = {arXiv preprint: http://arxiv.org/abs/2302.03133},
    doi = {10.48550/arXiv.2302.03133},
	publisher = {arXiv},
	author = {He, Huan and Queen, Owen and Koker, Teddy and Cuevas, Consuelo and Tsiligkaridis, Theodoros and Zitnik, Marinka},
	year = {2023},
}

@misc{Raghu2023,
	title = {Sequential {Multi}-{Dimensional} {Self}-{Supervised} {Learning} for {Clinical} {Time} {Series}},
    note = {arXiv preprint: http://arxiv.org/abs/2307.10923},
	doi = {10.48550/arXiv.2307.10923},
	publisher = {arXiv},
	author = {Raghu, Aniruddh and Chandak, Payal and Alam, Ridwan and Guttag, John and Stultz, Collin M.},
	year = {2023},
}

@misc{Jin2022,
	title = {Domain {Adaptation} for {Time} {Series} {Forecasting} via {Attention} {Sharing}},
    note = {arXiv preprint: http://arxiv.org/abs/2102.06828},
	doi = {10.48550/arXiv.2102.06828},
	publisher = {arXiv},
	author = {Jin, Xiaoyong and Park, Youngsuk and Maddix, Danielle C. and Wang, Hao and Wang, Yuyang},
	year = {2022},
}

@misc{Lai2023,
	title = {Context-aware {Domain} {Adaptation} for {Time} {Series} {Anomaly} {Detection}},
    note = {arXiv preprint: http://arxiv.org/abs/2304.07453},
	doi = {10.48550/arXiv.2304.07453},
	publisher = {arXiv},
	author = {Lai, Kwei-Herng and Wang, Lan and Chen, Huiyuan and Zhou, Kaixiong and Wang, Fei and Yang, Hao and Hu, Xia},
	year = {2023},
}

@book{Crainiceanu2006,
	address = {New York},
	edition = {2},
	title = {Measurement {Error} in {Nonlinear} {Models}: {A} {Modern} {Perspective}, {Second} {Edition}},
	shorttitle = {Measurement {Error} in {Nonlinear} {Models}},
	publisher = {Chapman and Hall/CRC},
	author = {Crainiceanu, David Ruppert and Leonard A. Stefanski and Ciprian M., Raymond J. Carroll},
	year = {2006},
	doi = {10.1201/9781420010138},
}

@article{Bauer2009,
	title = {Psychometric approaches for developing commensurate measures across independent studies: traditional and new models},
	volume = {14},
	issn = {1082-989X},
	doi = {10.1037/a0015583},
	number = {2},
	journal = {Psychological Methods},
	author = {Bauer, Daniel J. and Hussong, Andrea M.},
	year = {2009},
	pmid = {19485624},
	pmcid = {PMC2780030},
	pages = {101--125},
}

@article{Buuren2005,
	title = {Improving comparability of existing data by {Response} {Conversion}.},
	volume = {21},
	journal = {Journal of Official Statistics},
	author = {Van Buuren, Stef and Eyres, Sophie and Tennant, Alan and Hopman-Rock, Marijke},
	year = {2005},
	pages = {53--72},
}

@article{Heuvel2020,
	title = {Latent variable models for harmonization of test scores: {A} case study on memory},
	volume = {62},
	doi = {10.1002/bimj.201800146},
	number = {1},
	journal = {Biometrical Journal},
	author = {van den Heuvel, Edwin R. and Griffith, Lauren E. and Sohel, Nazmul and Fortier, Isabel and Muniz-Terrera, Graciela and Raina, Parminder},
	year = {2020},
	pages = {34--52},
}

@article{Bland1990,
	title = {A note on the use of the intraclass correlation coefficient in the evaluation of agreement between two methods of measurement},
	volume = {20},
	doi = {10.1016/0010-4825(90)90013-f},
	number = {5},
	journal = {Computers in Biology and Medicine},
	author = {Bland, J. M. and Altman, D. G.},
	year = {1990},
	pmid = {2257734},
	pages = {337--340},
}

@article{ProustLima2014,
	title = {Joint latent class models for longitudinal and time-to-event data: {A} review},
	volume = {23},
	doi = {10.1177/0962280212445839},
	number = {1},
	journal = {Statistical methods in medical research},
	author = {Proust-Lima, Cécile and Séne, Mbéry and Taylor, Jeremy MG and Jacqmin-Gadda, Hélène},
	year = {2014},
	pmid = {22517270},
	pmcid = {PMC5863083},
	pages = {74--90},
}

@article{ProustLima2023,
	title = {Describing complex disease progression using joint latent class models for multivariate longitudinal markers and clinical endpoints},
	volume = {42},
	doi = {10.1002/sim.9844},
	number = {22},
	journal = {Statistics in Medicine},
	author = {Proust-Lima, Cécile and Saulnier, Tiphaine and Philipps, Viviane and Traon, Anne Pavy-Le and Péran, Patrice and Rascol, Olivier and Meissner, Wassilios G. and Foubert-Samier, Alexandra},
	year = {2023},
	pages = {3996--4014},
}

@inproceedings{Goodfellow2014,
  title={Generative adversarial nets},
  author={Goodfellow, Ian and Pouget-Abadie, Jean and Mirza, Mehdi and Xu, Bing and Warde-Farley, David and Ozair, Sherjil and Courville, Aaron and Bengio, Yoshua},
  booktitle={Advances in neural information processing systems},
  volume={27},
  year={2014}
}

@article{Pechmann2019,
    title={SMArtCARE - A platform to collect real-life outcome data of patients with spinal muscular atrophy},
    author={Astrid Pechmann and Kirsten König and Günther Bernert and Kristina Schachtrup and Ulrike Schara and David Schorling and Inge Schwersenz and Sabine Stein and Adrian Tassoni and Sibylle Vogt and Maggie C. Walter and Hanns Lochmüller and Janbernd Kirschner},
    journal={Orphanet Journal of Rare Diseases},
    volume={14},
    number={18},
    year={2019}
}

@article{Schorling2020,
    author = {David C Schorling and Astrid Pechmann and Janbernd Kirschner},
    title = {Advances in Treatment of Spinal Muscular Atrophy - New Phenotypes, New Challenges, New Implications for Care},
    journal = {Journal of neuromuscular diseases},
    year = {2020},
    doi = {https://doi.org/10.3233/JND-190424},
    volume = {7},
    number = {1},
    pages = {1-13}
}

@article{Mazzone2017,
    title = {Revised upper limb module for spinal muscular atrophy: Development of a new module},
    author = {Elena S Mazzone and Anna Mayhew and Jacqueline Montes and Danielle Ramsey and Lavinia Fanelli and Sally Dunaway Young and Rachel Salazar and Roberto De Sanctis and Amy Pasternak and Allan Glanzman and Giorgia Coratti and Matthew Civitello and Nicola Forcina and Richard Gee and Tina Duong and Marika Pane and Mariacristina Scoto and Maria Carmela Pera and Sonia Messina and Gihan Tennekoon and John W Day and Basil T Darras and Darryl C De Vivo and Richard Finkel and Francesco Muntoni and Eugenio Mercuri},
    year = {2017},
    journal = {Muscle Nerve},
    volume = {55},
    number = {6},
    pages = {869-874},
    doi = {10.1002/mus.25430}
}

@article{OHagen2007,
	title = {An expanded version of the {Hammersmith} {Functional} {Motor} {Scale} for {SMA} {II} and {III} patients},
	volume = {17},
	issn = {0960-8966},
	doi = {10.1016/j.nmd.2007.05.009},
	number = {9-10},
	journal = {Neuromuscular disorders: NMD},
	author = {O'Hagen, Jessica M. and Glanzman, Allan M. and McDermott, Michael P. and Ryan, Patricia A. and Flickinger, Jean and Quigley, Janet and Riley, Susan and Sanborn, Erica and Irvine, Carrie and Martens, William B. and Annis, Christine and Tawil, Rabi and Oskoui, Maryam and Darras, Basil T. and Finkel, Richard S. and De Vivo, Darryl C.},
	year = {2007},
	pmid = {17658255},
	pages = {693--697},
}

@article{Messina2022,
	title = {Spinal muscular atrophy: state of the art and new therapeutic strategies},
	volume = {43},
	doi = {10.1007/s10072-021-05258-3},
	number = {Suppl 2},
	journal = {Neurological Sciences: Official Journal of the Italian Neurological Society and of the Italian Society of Clinical Neurophysiology},
	author = {Messina, Sonia and Sframeli, Maria and Maggi, Lorenzo and D'Amico, Adele and Bruno, Claudio and Comi, Giacomo and Mercuri, Eugenio},
	year = {2022},
	pmid = {33871750},
	pages = {615--624},
}

@article{Glanzman2011,
	title = {Validation of the {Expanded} {Hammersmith} {Functional} {Motor} {Scale} in spinal muscular atrophy type {II} and {III}},
	volume = {26},
	doi = {10.1177/0883073811420294},
	number = {12},
	journal = {Journal of Child Neurology},
	author = {Glanzman, Allan M. and O'Hagen, Jessica M. and McDermott, Michael P. and Martens, William B. and Flickinger, Jean and Riley, Susan and Quigley, Janet and Montes, Jacqueline and Dunaway, Sally and Deng, Liyong and Chung, Wendy K. and Tawil, Rabi and Darras, Basil T. and De Vivo, Darryl C. and Kaufmann, Petra and Finkel, Richard S. and {Pediatric Neuromuscular Clinical Research Network for Spinal Muscular Atrophy (PNCR)} and {Muscle Study Group (MSG)}},
	year = {2011},
	pmid = {21940700},
	pages = {1499--1507},
}

@article{Lefebvre1995,
	title = {Identification and characterization of a spinal muscular atrophy-determining gene},
	volume = {80},
	doi = {10.1016/0092-8674(95)90460-3},
	number = {1},
	journal = {Cell},
	author = {Lefebvre, S. and Bürglen, L. and Reboullet, S. and Clermont, O. and Burlet, P. and Viollet, L. and Benichou, B. and Cruaud, C. and Millasseau, P. and Zeviani, M.},
	year = {1995},
	pmid = {7813012},
	pages = {155--165},
}
}

\end{document}